\documentclass{article} %

\usepackage[preprint]{colm2026_conference}

\usepackage{microtype}
\usepackage{hyperref}
\usepackage{url}
\usepackage{booktabs}

\usepackage{lineno}

\definecolor{darkblue}{rgb}{0, 0, 0.5}
\hypersetup{colorlinks=true, citecolor=darkblue, linkcolor=darkblue, urlcolor=darkblue}

\usepackage[whole]{bxcjkjatype}
\usepackage{xspace}

\usepackage[T1]{fontenc}

\def\VersionWithComments{}

\definecolor{snsblue}{RGB}{59, 117, 175}
\definecolor{snsgreen}{RGB}{81, 158, 62}
\definecolor{snsorange}{RGB}{239, 134, 53}
\definecolor{snsred}{RGB}{234, 51, 35}
\definecolor{heatmapred}{RGB}{214, 38, 40}
\definecolor{heatmapblue}{HTML}{2077B4}

\ifdefined\VersionWithComments%
    \usepackage[colorinlistoftodos,textsize=footnotesize]{todonotes}
\else
    \usepackage[disable]{todonotes}
\fi

\setuptodonotes{inline}

\newlength{\needcitelength}
\newcommand{\needcite}[1]{\settowidth{\needcitelength}{cites: {#1}}\todo[noinlinepar,inlinewidth=\the\needcitelength,size=\scriptsize]{cites: {#1}}}
\newcommand{\needref}[1]{\settowidth{\needcitelength}{ref: {#1}}\todo[color=yellow,noinlinepar,inlinewidth=\the\needcitelength,size=\scriptsize]{ref: {#1}}}

\usepackage{amsmath}
\usepackage{amssymb}
\usepackage{amsfonts}
\usepackage{bbm}
\usepackage{bm}

\usepackage[capitalise,english,nameinlink]{cleveref} %

\crefname{figure}{\text{Fig.}}{\text{Fig.}}
\crefname{section}{\S}{\S\S}
\crefname{equation}{\text{Eq.}}{\text{Eq.}}
\crefname{table}{\text{Tbl.}}{\text{Tbl.}}
\creflabelformat{equation}{#2\textup{#1}#3}
\crefname{appendix}{\mbox{App.}}{\mbox{App.}}

\usepackage{cite}

\usepackage{booktabs}
\usepackage{bm}
\usepackage{tabularx}
\usepackage{arydshln}

\title{Exclusive Unlearning}

\author{
Mutsumi Sasaki${}^{1,2}$ \quad
Kouta Nakayama${}^{2}$ \\
\bf{
Yusuke Miyao${}^{3,2}$ \quad
Yohei Oseki${}^{3,2}$ \quad
Masaru Isonuma${}^{2,1,3}$ }
\\
${}^1$Tohoku University \quad
${}^2$NII LLMC \quad
${}^3$The University of Tokyo \\
\small{
\textbf{Correspondence:} \href{mailto:mutsumi.sasaki@dc.tohoku.ac.jp}{mutsumi.sasaki@dc.tohoku.ac.jp}
}
}

\begin{document}

\ifcolmsubmission
\linenumbers
\fi

\maketitle

\begin{abstract}
When introducing Large Language Models (LLMs) into industrial applications, such as healthcare and education, the risk of generating harmful content becomes a significant challenge.
While existing machine unlearning methods can erase specific harmful knowledge and expressions, diverse harmful content makes comprehensive removal difficult.
In this study, instead of individually listing targets for forgetting, we propose Exclusive Unlearning (EU), which aims for broad harm removal by extensively forgetting everything except for the knowledge and expressions we wish to retain.
We demonstrate that through Exclusive Unlearning, it is possible to obtain a model that ensures safety against a wide range of inputs, including jailbreaks, while maintaining the ability to respond to diverse instructions related to specific domains such as medicine and mathematics.
\footnote{
\includegraphics[width=1em]{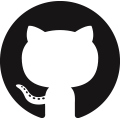}
\href{https://github.com/cl-tohoku/ExclusiveUnlearning}{github.com/cl-tohoku/ExclusiveUnlearning}
}

\end{abstract}

\section{Introduction}
\label{sec:introduction}
Large Language Models (LLMs) inherently possess risks such as the leakage of private information and copyright infringement~\citep{carlini2023quantifying, dou-etal-2025-avoiding, huang-etal-2022-large,ippolito-etal-2023-preventing}.
Because LLMs are trained on massive corpora, it is unrealistic to remove all harmful content from the training data in advance.
Consequently, unlearning~\citep{liu2025rethinking, liu-etal-2024-towards-safer,si2023knowledgeunlearningllmstasks, yuan2025a, wang2025large}, which removes specific knowledge from a trained LLM, has attracted attention as a low-cost alternative.

However, traditional unlearning research is primarily based on the problem definition of ``what to forget'', which assumes an exhaustive enumeration of the harmful expressions to be erased. In practice, it is difficult to prepare diverse unlearning datasets that anticipate every possible attack pattern, and generalization to unseen harmful questions not included during training remains insufficient~\citep{yao-etal-2024-machine,zhang2025from}.
In particular, the recent diversification of jailbreak attack methods~\citep{liu2024jailbreakingchatgptpromptengineering,shen2024donowcharacterizingevaluating,yu2024gptfuzzerredteaminglarge,NEURIPS2023_fd661313,zou2023universaltransferableadversarialattacks,10579515,liu2024autodan} makes it increasingly difficult to defend against every input designed to elicit harmful generation simply by specifying unlearning data.
While recent studies~\citep{lu2024eraserjailbreakingdefenselarge,liu-etal-2024-towards-safer} have proposed methods to generalize to unseen harmful or jailbreak inputs using representative harmful data, such generalization is still inadequate and the approach is becoming increasingly complex~\citep{li2025safellmunlearningharmfuloutputs}.

\begin{figure*}[t!]
    \centering
    \includegraphics[width=\linewidth]{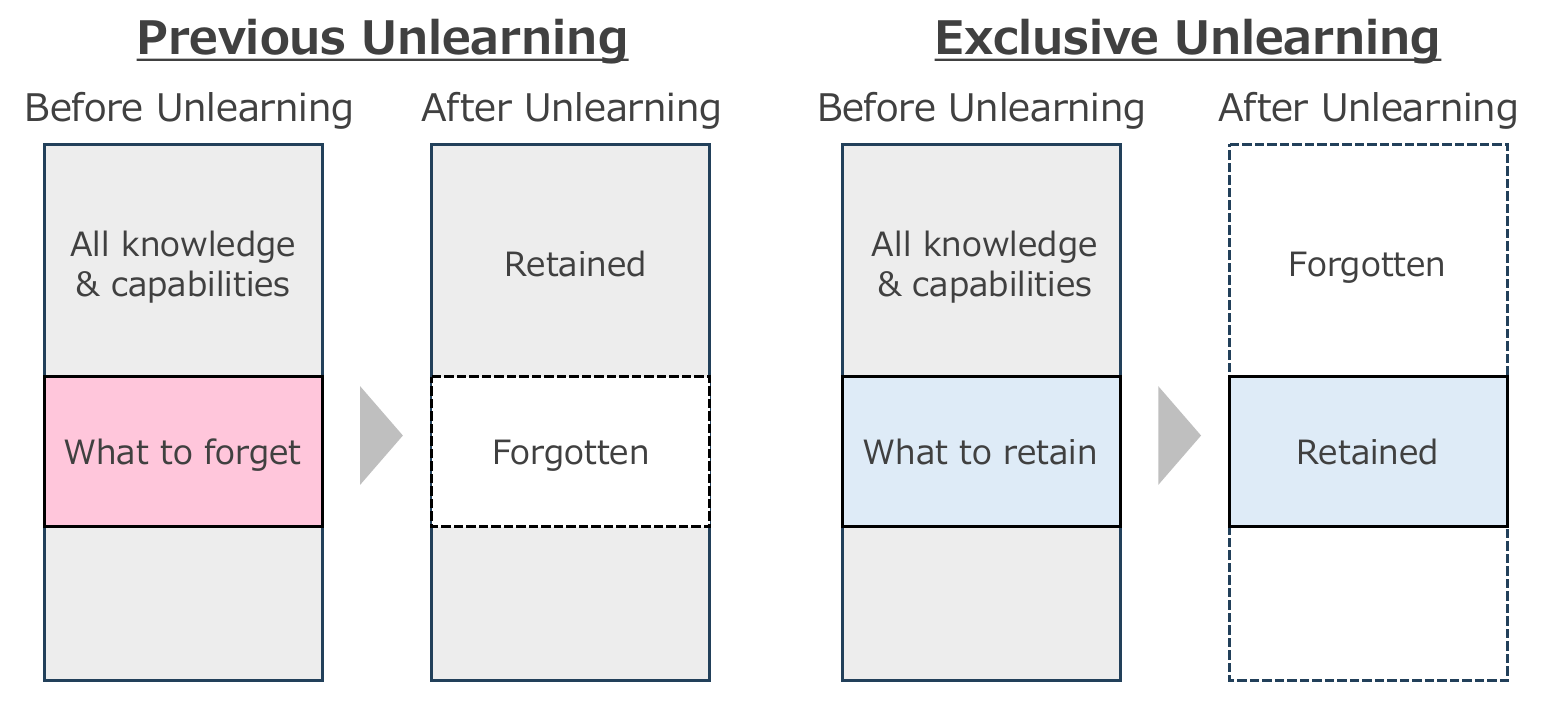}
    \caption{
    \textbf{Overview of Exclusive Unlearning (EU).}
    By specifying only ``what to retain'' and exclusively forgetting everything else, this method can circumvent the limitations of traditional unlearning, where generalizing to unseen harmful domains not included in the unlearning data is difficult.
    }
    \label{fig:figure1}
\end{figure*}

In this study, we attempt to solve this issue by shifting the problem definition of unlearning through a simple yet effective approach.
Specifically, instead of specifying ``what to forget'', we propose Exclusive Unlearning (EU), where we specify only ``what to retain'' through retention data and uniformly forget everything else (\cref{fig:figure1}).
As a concrete approach, we sample diverse text from the model itself and bring the model's generation probabilities closer to a uniform distribution to induce forgetting, while simultaneously performing standard fine-tuning on the target dataset to be retained.
This allows the model to exclusively unlearn everything except for the specific abilities required to answer instructions and tasks defined by the retention dataset. 
While traditional unlearning requires specific forgetting data, our proposed method does not require the item-by-item enumeration of harmful behaviors, thereby fundamentally bypassing the limitations of data-dependent unlearning.
Although this setting is less suited for general-purpose LLMs because it requires a clear definition of target abilities, it is ideal for domain-specific LLM application scenarios, such as medical diagnostic support or educational assistance.
For instance, in a medical application, the retention target can be clearly defined as medical knowledge and instruction-following capability for medical queries; in such cases, our method can achieve defensive performance that far exceeds conventional approaches.

In our experiments, we demonstrated the effectiveness of this method using instruction tuning datasets for medicine and mathematics, looking toward applications in the healthcare and education sectors. 
After Exclusive Unlearning, the model maintained its target capabilities at a level equal to or higher than the original model, while showing high defensive performance against a wide range of harmful inputs, including jailbreaks. 
In particular, in terms of generalization to diverse harmful inputs, our results significantly outperformed existing methods, including safety alignment via Direct Preference Optimization (DPO), standard unlearning based on harmful data enumeration, and prior unlearning methods~\citep{lu2024eraserjailbreakingdefenselarge,liu-etal-2024-towards-safer} that build on such harmful-data-based unlearning with additional mechanisms to improve forgetting generalization to unseen harmful inputs.

The primary contributions of this paper are as follows:

(1) We propose a new retention-only framework that fundamentally outperforms the conventional paradigm of enumerating and generalizing from harmful data in terms of forgetting unseen and jailbreak inputs, albeit within a specific application scope.

(2) We implement this framework with a simple approach by changing the focus of existing unlearning, and empirically demonstrate its superior effectiveness.

\section{Methodology}
\label{sec:methodology}
Our objective is to retain only the abilities required to solve the tasks in a target dataset, while effectively forgetting all other inputs.
In unlearning, parameter optimization is generally performed by combining a forgetting loss $\mathcal{L}_{\text{forget}}(\theta)$ and a retention loss $\mathcal{L}_{\text{retain}}(\theta)$.
The most fundamental approach in conventional unlearning is Gradient Ascent (GA), which sets $\mathcal{L}_{\text{forget}}(\theta)$ to minimize the log-likelihood of a pre-defined forget dataset $\mathcal{D}_{\text{forget}}$
\begin{equation}
\mathcal{L}_{\text{forget}}(\theta) =
\mathbb{E}_{x \sim \mathcal{D}_{\text{forget}}}\left[\log p_\theta(x) \right].
\label{eq:gradient_ascent}
\end{equation}
However, performing unlearning on a fixed $\mathcal{D}_{\text{forget}}$ makes it difficult to generalize to a broad range of harmful inputs that lie outside the scope of that specific dataset.

To address this, we propose a method that requires no explicit forget dataset by specifying only the target dataset to be retained and exclusively forgetting non-retained knowledge and capabilities. 
A key feature of our approach is the unlearning of text generated by the model itself:
\begin{equation}
\mathcal{L}_{\text{forget}}(\theta) =
\mathbb{E}_{x \sim p_\theta}\left[\log p_\theta(x) \right].
\label{eq:forget}
\end{equation}
This design is based on the intuition that the model's internal knowledge and capabilities are represented in its self-generated outputs.
Actually, this objective corresponds to maximizing the entropy of the model's predictions. 
Maximizing the entropy is equivalent to minimizing the Kullback-Leibler divergence between the model's prediction and a uniform distribution $p_u$:
\begin{equation}
\mathbb{E}_{x \sim p_\theta}\left[\log p_\theta(x) \right] = \mathrm{KL}\!\left[p_\theta \,\|\, p_u\right] - T\log V
\label{eq:forget_kl}
\end{equation}
where $V$ denotes the vocabulary size, and $T$ is the sequence length of $x$. 
As a result, the model's predictions are forced into a uniform distribution, ensuring that all non-target knowledge is thoroughly unlearned.

This approach resolves the fundamental challenge in existing unlearning methods: the difficulty of anticipating and curating data for every possible harmful query or attack pattern.
While this results in the removal of knowledge and capabilities regardless of their harmfulness, our method is highly effective for domain-specific applications, where the capabilities to be retained are clearly defined.

For the retention objective, we minimize the negative log-likelihood over the retention dataset $\mathcal{D}_{\text{retain}}$, following standard fine-tuning procedures:
\begin{equation}
\mathcal{L}_{\text{retain}}(\theta) = \mathbb{E}_{x \sim \mathcal{D}_{\text{retain}}}\left[ - \log p_\theta(x) \right].
\label{eq:retain}
\end{equation}
Finally, we integrate \cref{eq:forget} and \cref{eq:retain} to achieve the simultaneous retention of specific expertise and the exclusive unlearning of all other information:
\begin{equation}
\mathcal{L}(\theta) = \lambda \cdot \mathcal{L}_{\text{forget}}(\theta) + (1-\lambda) \cdot \mathcal{L}_{\text{retain}}(\theta),
\label{eq:overall}
\end{equation}
where the balancing parameter $\lambda$ controls the trade-off between forgetting and retention.

\section{Experimental Setup}
\subsection{Training Setup}
Assuming applications in the medical and mathematical domains, we use two instruction tuning datasets as retention targets: MedInstruct-52k from AlpaCare~\citep{zhang2025alpacareinstructiontunedlargelanguagemodels} for the medical domain, and MetaMathQA from MetaMath~\citep{yu2024metamath} for the mathematical domain. 

For the unlearning process, we utilize text generated by the model itself, sampling with a batch size of 4 over 10,000 steps.
For the base models, we use Llama-3.2-1B-Instruct, Llama-3.2-3B-Instruct~\citep{meta2024llama32}, OLMo-2-1B-Instruct, and OLMo-2-7B-Instruct~\citep{olmo20252olmo2furious}. 
Full details of the self-generation procedure and hyperparameter analysis are provided in \cref{sec:training_details}.

We also tune the balancing parameter $\lambda$ over a set of candidate values and report the best-performing setting.
The sensitivity and stability with respect to $\lambda$ are discussed in \cref{sec:lambda_sensitivity}.

\subsection{Evaluation Setup}
\label{sec:evaluation_setup}
In the medical domain, we conduct both benchmark and LLM-based evaluations. 
For benchmark evaluation, we use four medical multiple-choice QA datasets (MedQA~\citep{jin2020diseasedoespatienthave}, HeadQA~\citep{vilares-gomez-rodriguez-2019-head}, PubmedQA~\citep{jin-etal-2019-pubmedqa}, and MEDMCQA~\citep{pmlr-v174-pal22a}) and the summarization dataset (MeQSUM~\citep{ben-abacha-demner-fushman-2019-summarization}), reporting accuracy for QA and ROUGE-L for summarization following the lm-evaluation-harness~\citep{eval-harness}. 
For LLM-based evaluation, we use MedInstruct-test (216 medical instructions) and iCliniq~\citep{li2023chatdoctormedicalchatmodel} (1,000 patient queries), and use GPT-5-2025-08-07~\citep{singh2025openaigpt5card} to score responses on four five-point dimensions: Accurate, Useful, Comprehensible, and Succinct, adapted from PDSQI-9~\citep{croxford2025evaluating}. 
Details are provided in \cref{sec:appendix_medical_llm_eval}.

In the mathematical domain, we evaluate on GSM8K~\citep{cobbe2021trainingverifierssolvemath}, MATH~\citep{hendrycks2021measuring}, and MathQA~\citep{amini-etal-2019-mathqa}, reporting Exact Match (EM) for generation tasks and accuracy for multiple-choice QA following the lm-evaluation-harness.

To evaluate forgetting performance, we measure defensive performance against harmful and jailbreak inputs.
We use two harmful-question sets, \textbf{Harm-1} and \textbf{Harm-2}, constructed from GPTFUZZER~\citep{yu2024gptfuzzerredteaminglarge} and WildAttack~\citep{shen2024donowcharacterizingevaluating}, respectively. We further construct two corresponding jailbreak sets, \textbf{JB-1} and \textbf{JB-2}, by combining them with 20 types of unseen jailbreak prompts from prior work~\citep{zhang2025from}. Details are provided in \cref{sec:detail_of_jailbreak}.
Following prior work~\citep{zhang2025from}, we use ShieldLM-14B-qwen~\citep{zhang2024shieldlm} as a safety judge and report Attack Success Rate (ASR).

In our unlearning objective, based on \cref{eq:forget_kl}, forgotten inputs are expected to induce output distributions close to uniform. 
We therefore evaluate whether a given input has been forgotten according to whether its output distribution is sufficiently close to uniform (determined by a threshold $\Delta(x)$; see \cref{sec:appendix_delta} for details). 
For inputs judged to be forgotten, our method returns a fixed refusal response, ``I can't answer the instruction.'', thereby avoiding incoherent outputs and ensuring the safety of the response for those inputs.

\subsection{Baselines for Forgetting and Retention}
\label{sec:baseline}
We compare our method with both retention and forgetting baselines. 
For retention, we use the original model before unlearning and a model trained only with the retention objective in \cref{eq:retain}, denoted as \textbf{RetainOnlyFT}. 

For forgetting, we consider four baselines. 
\textbf{DPO} applies Direct Preference Optimization based on PKU-SafeRLHF~\citep{ji-etal-2025-pku} together with the retention objective. 
\textbf{Unlearning} optimizes the forgetting objective in \cref{eq:gradient_ascent} on 40{,}000 harmful examples from PKU-SafeRLHF-QA together with the same retention objective. 
We further adapt two prior unlearning methods designed to improve generalization to jailbreak-style harmful inputs: \textbf{Eraser}~\citep{lu2024eraserjailbreakingdefenselarge}, which applies forgetting to perturbed harmful inputs with random prefix/suffix noise, and \textbf{SKU}~\citep{liu-etal-2024-towards-safer}, which subtracts a harmful-behavior update direction at the parameter level. 
For fair comparison, DPO uses 40{,}000 preference pairs, and all harmful-data-based baselines use the same 40{,}000 PKU-SafeRLHF-QA examples and the same target-domain retention data. 
In addition, \textbf{Unlearning} and \textbf{Eraser} also produce uniform output distributions on forgotten inputs. 
Therefore, we apply the forgetting criterion described in \cref{sec:appendix_delta} to them as well, and return a fixed response for outputs judged as forgotten.
Details of each baseline are provided in \cref{sec:baselines_appendix}.

\begin{table*}[t]
    \centering
    \small
    \setlength{\tabcolsep}{4pt}
    \renewcommand{\arraystretch}{1.06}
    \begin{tabular}{l|cc|cccc}
    \toprule
    & \multicolumn{2}{c|}{\textbf{Retention}$\uparrow$}
    & \multicolumn{4}{c}{\textbf{Forgetting (ASR)}$\downarrow$} \\
    \cmidrule(lr){2-3}\cmidrule(lr){4-7}
    \textbf{Setting}
    & \textbf{MC Avg.}
    & \textbf{MeQSum}
    & \textbf{Harm-1}
    & \textbf{Harm-2}
    & \textbf{JB-1}
    & \textbf{JB-2} \\
    \midrule

    \textbf{Llama-3.2-1B-Instruct} & 0.3672 & 0.1192 & 7.0 & 8.8 & 52.5 & 51.1 \\
    RetainOnlyFT                  & 0.3820 & 0.2582 & 16.0 & 19.4 & 48.6 & 52.6 \\
    DPO           & 0.3712 & 0.2477 & 2.0 & 2.8 & 6.7 & 5.3 \\
    Unlearning ($\lambda=0.6$)           & 0.3898 & 0.2639 & 0.0 & 0.0 & 21.0 & 22.9 \\
    Eraser ($\lambda=0.6$)            & 0.3720 & 0.1754 & 0.0 & 0.5 & 17.2 & 16.0 \\
    SKU           & 0.3314 & 0.1122 & 5.0 & 6.0 & 11.9 & 10.9 \\
    EU ($\lambda=0.6$)            & 0.3605 & 0.2397 & 0.0 & 0.0 & 0.0 & 0.3 \\
    \midrule

    \textbf{Llama-3.2-3B-Instruct} & 0.4290 & 0.0863 & 2.0 & 4.6 & 18.1 & 10.2 \\
    RetainOnlyFT                   & 0.4496 & 0.2394 & 9.0 & 18.9 & 60.3 & 58.9 \\
    DPO          & 0.5009 & 0.1504 & 2.0 & 2.3 & 3.5 & 2.5 \\
    Unlearning ($\lambda=0.6$)           & 0.4695 & 0.2083 & 0.0 & 0.0 & 25.5 & 23.3 \\
    Eraser ($\lambda=0.6$)            & 0.4500 & 0.1919 & 0.0 & 0.0 & 0.7 & 0.3 \\
    SKU           & 0.4198 & 0.1079 & 0.0 & 1.3 & 7.6 & 5.7 \\
    EU ($\lambda=0.6$)             & 0.4468 & 0.2015 & 0.0 & 0.0 & 0.1 & 0.2 \\
    \midrule

    \textbf{OLMo-2-1B-Instruct} & 0.3386 & 0.1475 & 1.0 & 0.9 & 14.2 & 11.1 \\
    RetainOnlyFT                     & 0.3567 & 0.2832 & 11.0 & 18.9 & 39.2 & 52.6 \\
    DPO           & 0.3507 & 0.1524 & 0.0 & 0.1 & 20.8 & 20.1 \\
    Unlearning ($\lambda=0.95$)           & 0.3559 & 0.2756 & 0.0 & 0.0 & 22.5 & 22.9 \\
    Eraser ($\lambda=0.95$)            & 0.3764 & 0.2417 & 1.0 & 9.2 & 7.3 & 6.5 \\
    SKU           & 0.3713 & 0.1079 & 0.0 & 0.0 & 6.2 & 5.7 \\
    EU ($\lambda=0.95$)              & 0.3541 & 0.2944 & 1.0 & 0.0 & 0.0 & 0.0 \\
    \midrule

    \textbf{OLMo-2-7B-Instruct} & 0.5063 & 0.2449 & 0.0 & 0.9 & 29.2 & 26.6 \\
    RetainOnlyFT                     & 0.4800 & 0.2587 & 6.0 & 13.4 & 45.9 & 45.6 \\
    DPO           & 0.4880 & 0.1841 & 1.0 & 0.0 & 28.6 & 25.8 \\
    Unlearning ($\lambda=0.6$)           & 0.4633 & 0.2367 & 0.0 & 0.0 & 42.1 & 41.9 \\
    Eraser ($\lambda=0.6$)            & 0.4582 & 0.2398 & 0.0 & 0.0 & 45.1 & 40.3 \\
    SKU           & 0.4817 & 0.1333 & 0.0 & 0.0 & 7.0 & 6.5 \\
    EU ($\lambda=0.6$)               & 0.4753 & 0.2868 & 0.0 & 0.0 & 0.3 & 0.2 \\

    \bottomrule
    \end{tabular}
    \caption{
Retention and forgetting performance for medical-domain experiments.
The left block reports retention performance, where we summarize the four medical multiple-choice benchmarks by their average (MC Avg.), while reporting MeQSUM separately as the summarization task.
Higher is better for retention.
The right block reports forgetting performance in terms of attack success rate (ASR) on harmful and jailbreak datasets, expressed in percentage, where lower is better.
}
\label{tab:medical_retain_forget_compact}
\end{table*}

\begin{table*}[t]
    \centering
    \footnotesize
    \setlength{\tabcolsep}{3.5pt}
    \renewcommand{\arraystretch}{1.03}
    \resizebox{\textwidth}{!}{%
    \begin{tabular}{lccccc|ccccc}
    \toprule
    & \multicolumn{5}{c|}{\textbf{MedInstruct-test}} 
    & \multicolumn{5}{c}{\textbf{iCliniq}} \\
    \cmidrule(lr){2-6} \cmidrule(lr){7-11}
    \textbf{Setting}
    & \textbf{Acc.}
    & \textbf{Use.}
    & \textbf{Comp.}
    & \textbf{Suc.}
    & \textbf{Avg}
    & \textbf{Acc.}
    & \textbf{Use.}
    & \textbf{Comp.}
    & \textbf{Suc.}
    & \textbf{Avg} \\
    \midrule

    \textbf{Llama-3.2-1B-Instruct}
    & 2.65 & 2.73 & 4.84 & 2.82 & 3.26
    & 2.64 & 2.03 & 4.77 & 3.10 & 3.14 \\
    RetainOnlyFT
    & 3.68 & 2.97 & 4.56 & 3.94 & 3.79
    & 2.84 & 2.20 & 4.71 & 4.00 & 3.44 \\
    EU ($\lambda=0.6$)
    & 3.62 & 2.93 & 4.52 & 4.03 & 3.78
    & 2.90 & 2.22 & 4.74 & 3.97 & 3.46 \\
    \midrule

    \textbf{Llama-3.2-3B-Instruct}
    & 3.31 & 3.31 & 4.93 & 3.21 & 3.69
    & 2.99 & 2.56 & 4.90 & 3.43 & 3.47 \\
    RetainOnlyFT
    & 4.10 & 3.13 & 4.63 & 4.14 & 4.00
    & 3.32 & 2.48 & 4.80 & 3.98 & 3.65 \\
    EU ($\lambda=0.6$)
    & 4.14 & 3.14 & 4.60 & 4.07 & 3.99
    & 3.35 & 2.48 & 4.82 & 3.99 & 3.66 \\
    \midrule

    \textbf{OLMo-2-1B-Instruct}
    & 3.06 & 2.96 & 4.86 & 3.19 & 3.52
    & 2.51 & 2.36 & 4.70 & 2.82 & 3.10 \\
    RetainOnlyFT
    & 3.77 & 2.94 & 4.61 & 3.84 & 3.79
    & 3.00 & 2.31 & 4.74 & 3.89 & 3.49 \\
    EU ($\lambda=0.95$)
    & 3.86 & 2.91 & 4.66 & 3.95 & 3.85
    & 3.07 & 2.32 & 4.76 & 3.89 & 3.51 \\
    \midrule

    \textbf{OLMo-2-7B-Instruct}
    & 4.18 & 3.70 & 4.99 & 3.80 & 4.17
    & 3.64 & 3.08 & 4.95 & 3.65 & 3.83 \\
    RetainOnlyFT
    & 4.19 & 3.53 & 4.95 & 3.96 & 4.16
    & 3.60 & 2.63 & 4.79 & 4.09 & 3.78 \\
    EU ($\lambda=0.6$)
    & 4.26 & 3.31 & 4.82 & 4.09 & 4.12
    & 3.63 & 2.65 & 4.80 & 4.08 & 3.79 \\
    \bottomrule
    \end{tabular}%
    }
    \caption{
    Judge-based generation evaluation results on MedInstruct-test and iCliniq. 
    \textbf{Acc.}, \textbf{Use.}, \textbf{Comp.}, and \textbf{Suc.} denote \textbf{Accuracy}, \textbf{Usefulness}, \textbf{Comprehensibility}, and \textbf{Succinctness}, respectively.
    \textbf{Avg} denotes their average score.
    }
    \label{tab:retain_medinstruct_icliniq_gpt5}
\end{table*}

\section{Results}
\subsection{Overall Retention and Forgetting Performance}
\label{sec:main_results}
In the medical-retention setting, we first examine the medical benchmark results shown in \cref{tab:medical_retain_forget_compact}. 
For the base models, the ASR on baseline harmful questions is already relatively low, reflecting the fact that the Instruct models used in our experiments have already undergone safety alignment.  
However, the ASR on jailbreak inputs remains high, indicating that the original alignment alone is insufficient against adversarial prompt reformulations.  
When these models are further specialized to the medical domain via RetainOnlyFT, retention performance improves, but safety degrades substantially. In particular, ASR increases not only on jailbreak evaluations but also on the baseline harmful-question sets.  
In contrast, EU preserves competitive medical retention performance while substantially reducing ASR on both harmful and jailbreak inputs.  
That is, while reducing ASR to at most a few percent, EU maintains benchmark performance close to both the base model and RetainOnlyFT.

Next, \cref{tab:retain_medinstruct_icliniq_gpt5} shows the LLM-based evaluation results on free-form responses for MedInstruct-test and iCliniq along four dimensions.  
MedInstruct-test is a clinician-crafted free-form medical instruction evaluation set corresponding to MedInstruct-52k, which we used for retention. 
Across all models and all evaluation dimensions, EU achieves performance comparable to RetainOnlyFT, confirming that it preserves the ability to respond to diverse instructions within the retained domain.  
ICliniq is a separate medical dataset of real patient queries and is less directly aligned with the retention data format.  
EU remains comparable to RetainOnlyFT there across all evaluation dimensions as well, indicating that the retained generation capability extends beyond MedInstruct-style instructions to another external in-domain dataset.

Finally, the results under the setting of retaining mathematical capability, shown in \cref{sec:math_domain,tab:math_retain_forget_combined}, exhibit the same overall trend as in the medical domain.
Across all four models, EU reduces ASR on harmful and jailbreak inputs to nearly zero, while maintaining mathematical performance on both generative tasks (GSM8K and MATH) and the multiple-choice task (MathQA) at a level above the base model and roughly comparable to RetainOnlyFT.
This suggests that our method can be applied under settings that retain capabilities for a variety of tasks.

\subsection{Comparison with Baselines}
In \cref{tab:medical_retain_forget_compact} and \cref{tab:math_retain_forget_combined}, we compare EU with two baselines designed to improve safety against harmful and jailbreak inputs.

The first baseline, \textbf{DPO}, learns refusal behavior through safety alignment.  
Compared with the original base models, DPO substantially reduces the attack success rate (ASR), typically to around 5\%--10\%.  
However, EU further reduces ASR to nearly 0\%, especially on jailbreak inputs, indicating stronger robustness.  
This suggests that alignment-based suppression of harmful behavior is still insufficient for robust defense.  
At the same time, EU is trained under a stricter constraint than DPO, since it is designed to forget not only harmful behavior but also non-target benign capabilities outside the retention domain.  
Thus, our point is not that DPO is uniformly inferior, but that EU is more effective specifically for forgetting and defense against adversarial harmful inputs.

The second baseline, \textbf{Unlearning}, explicitly enumerates harmful data and applies unlearning to those examples.  
While this baseline reduces the ASR on the curated harmful-question sets themselves to nearly 0\%, its ASR on jailbreak inputs remains substantially higher than that of EU.  
Moreover, the Eraser and SKU baselines, which are based on prior work that also explicitly enumerates harmful data but aims to generalize to a broader range of harmful inputs, sometimes reduce the ASR on jailbreak inputs to nearly 0\% for particular task settings and models; however, overall, they still record ASRs ranging from around 5\% to several tens of percent.  
These results support our hypothesis that unlearning approaches which aim to defend against a broad range of attacks by generalizing from explicitly listed harmful examples do not generalize sufficiently to jailbreak attacks, and they suggest that EU, which broadly forgets everything outside the retention domain, is effective for comprehensive forgetting against harmful or jailbreak inputs.

\subsection{Sensitivity to the Hyperparameter Balancing Forgetting and Retention Loss}
\label{sec:lambda_sensitivity}
Next, we analyze the sensitivity of Exclusive Unlearning (EU) to the balance parameter $\lambda$ in \cref{eq:overall}.  
The sensitivity results for the medical-domain retention setting are shown in \cref{tab:medical_retain_forget_appendix}, and those for the mathematical-domain retention setting are shown in \cref{tab:math_retain_forget_appendix}.  
Overall, the results indicate that EU is robust to the choice of $\lambda$ on both the retention and forgetting sides, rather than relying on a single narrowly tuned setting.

From the perspective of retention, EU maintains stable target-domain performance across a broad range of $\lambda$ values in both the medical and mathematical domains.  
From the perspective of forgetting, the same robustness is observed.  
Across both domains, harmful and jailbreak ASR is consistently reduced to very low levels over a wide range of moderate $\lambda$ values.  
This indicates that the forgetting behavior of EU is stable and reproducible with respect to $\lambda$.

\section{Discussion}
\subsection{Understanding the Behavior of Exclusive Unlearning}
\label{sec:behavior_analysis}
To better understand the behavior of Exclusive Unlearning, we evaluate the EU-trained model (medical retention setting) on a broad set of general-domain benchmarks unrelated to either the medical domain or harmful inputs: ARC-Easy, ARC-Challenge~\citep{boratko-etal-2018-systematic}, PIQA~\citep{Bisk2020}, HellaSwag~\citep{zellers-etal-2019-hellaswag}, GSM8K~\citep{cobbe2021trainingverifierssolvemath}, and MMLU~\citep{hendrycks2021measuringmmlu}. 
The results are shown in 
\cref{tab:medical_retain_general_benign}. 
While performance drops across most tasks after EU, it does not fall entirely to chance level, indicating partial retention of some general-domain inputs.

\begin{figure*}[t!]
    \centering
    \includegraphics[width=\linewidth]{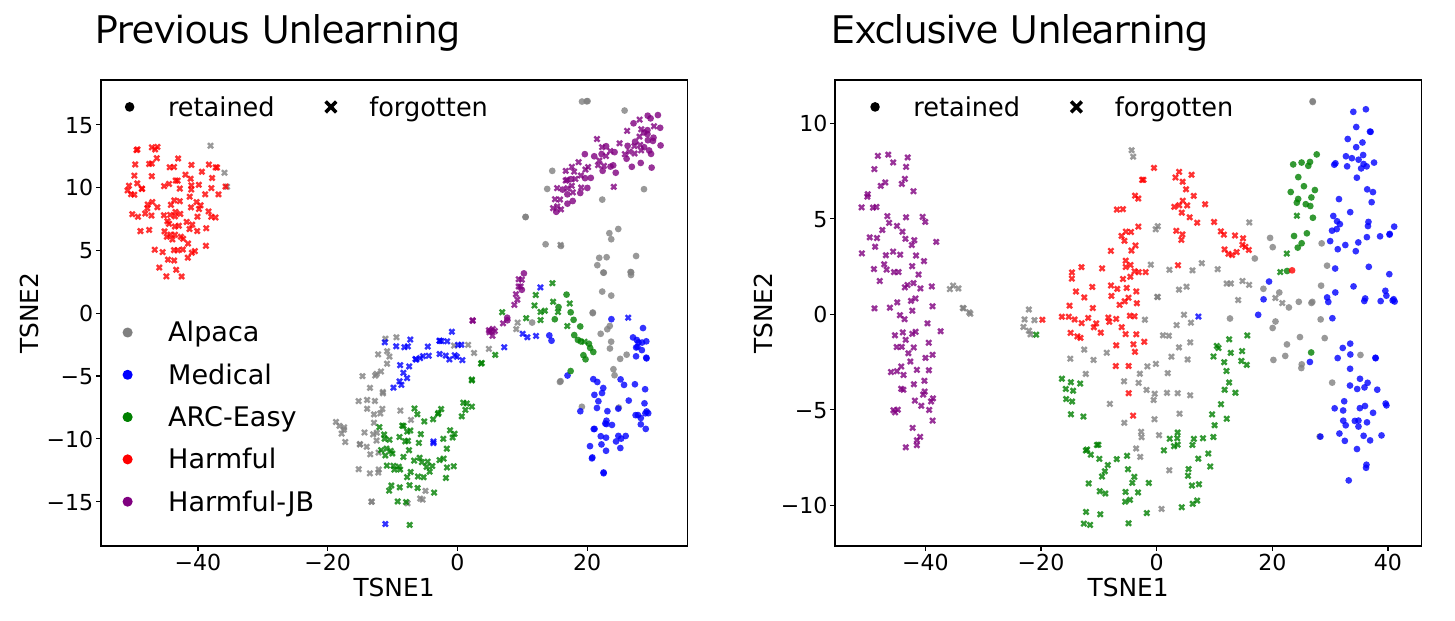}
    \caption{
t-SNE visualization of the final-layer representation space at the last-token position in the medical-retention setting of Llama-3.2-1B-Instruct.
We compare a conventional unlearning baseline (left) with Exclusive Unlearning (right) using five types of inputs: Alpaca, Medical, ARC-Easy, Harmful, and Harmful-JB.
Marker shape indicates whether the input is judged as retained or forgotten based on \cref{sec:appendix_delta}.
    }
    \label{fig:tsne_visualization}
\end{figure*}

To understand why harmful inputs are nonetheless forgotten effectively (as shown in \cref{tab:medical_retain_forget_compact}), we analyze the representation space of the EU-trained Llama-3.2-1B-Instruct. 
We feed four categories of inputs into the model: harmful inputs (100 samples from \textbf{Harm-1}), their jailbreak variants, medical data (100 samples from MedInstruct-test), and general-domain data unrelated to either (100 samples 
each from Alpaca and ARC-Easy). 
We apply t-SNE to the final-layer hidden states at the last token position and, using the criterion from \cref{sec:appendix_delta}, distinguish retained from forgotten inputs in the 
visualization. 
We perform the same analysis on the conventional unlearning baseline for comparison.
The results of Llama-3.2-1B-Instruct are shown in \cref{fig:tsne_visualization}. 
We observe the same overall tendency for other models as well; details are provided in \cref{sec:behavior_appendix}.

The visualization reveals the behavior underlying EU. 
In the EU model (right), some general-domain inputs lie close to the medical domain in representation space and are thus partially retained by the retention objective, which suggests why general-domain performance does not collapse to chance level.
Actually, among the general-domain inputs that were not judged as forgotten, we observed many examples that are semantically close to the retained medical domain, including instructions related to health and care in Alpaca and science questions pertaining to biology or health in ARC-Easy (see ~\cref{tab:retained_general_examples}). 
Harmful inputs, however, are represented far from the 
medical domain and are therefore consistently forgotten, explaining the strong defensive performance in \cref{tab:medical_retain_forget_compact}.

The contrast with conventional unlearning (left) is particularly informative for jailbreak robustness. 
Conventional unlearning targets the harmful region directly, achieving strong forgetting within that region. However, jailbreak prompts shift the representation of harmful inputs toward the general-domain region, causing some of them to escape forgetting. 
EU avoids this failure 
mode: since it forgets everything outside the retention domain, jailbreak inputs that are representationally distant from the medical domain are forgotten just as reliably as plain harmful inputs.

\begin{table*}[t]
    \centering
    \small
    \footnotesize
    \setlength{\tabcolsep}{4pt}
    \renewcommand{\arraystretch}{1.03}
    \begin{tabular}{lcccccc}
    \toprule
    \textbf{Setting}
    & \textbf{ARC-Easy}
    & \textbf{ARC-Challenge}
    & \textbf{PIQA}
    & \textbf{HellaSwag}
    & \textbf{GSM8K}
    & \textbf{MMLU} \\
    \midrule
    Metric
    & 4-choice
    & 4-choice
    & 2-choice
    & 4-choice
    & EM
    & 4-choice \\
    \midrule

    \textbf{Llama-3.2-1B-Instruct}
    & 0.5859 & 0.3072 & 0.7404 & 0.4498 & 0.1312 & 0.3126 \\
    EU ($\lambda=0.6$)
    & 0.4112 & 0.3072 & 0.5838 & 0.2656 & 0.0050 & 0.2489 \\
    \midrule

    \textbf{Llama-3.2-3B-Instruct}
    & 0.7045 & 0.3942 & 0.7579 & 0.5073 & 0.1850 & 0.3930 \\
    EU ($\lambda=0.6$)
    & 0.5741 & 0.3430 & 0.6066 & 0.2945 & 0.0000 & 0.2975 \\
    \midrule

    \textbf{OLMo-2-1B-Instruct}
    & 0.6915 & 0.3788 & 0.7476 & 0.5141 & 0.4268 & 0.3453 \\
    EU ($\lambda=0.95$)
    & 0.5320 & 0.2867 & 0.6306 & 0.3968 & 0.0106 & 0.2396 \\
    \midrule

    \textbf{OLMo-2-7B-Instruct}
    & 0.7630 & 0.4616 & 0.7731 & 0.5962 & 0.7127 & 0.4418 \\
    EU ($\lambda=0.6$)
    & 0.4785 & 0.3012 & 0.6415 & 0.3542 & 0.0037 & 0.2307 \\
    \bottomrule
    \end{tabular}
    \caption{
    General benchmark performance under the setting of retaining medical capability.
    }
    \label{tab:medical_retain_general_benign}
\end{table*}

\subsection{Design Analysis of Self-generated Text in Exclusive Unlearning}
A key design choice in EU is to use self-generated text as the source of forgetting, rather than relying on an explicitly curated forget corpus. 
To examine the validity of this design, we first compare self-generated text with two alternative sources under otherwise identical training configurations: one sampled from the Wikipedia subset of the Dolma corpus (Wiki), and another sampled from the entire Pile corpus (Pile). 
As shown in \cref{sec:appendix_self_generate,fig:asr_ablation}, the setting using the model's self-generated text achieves the strongest forgetting performance. 
This result supports the view that self-generated text provides an effective forgetting source for EU, likely because it more directly reflects the model's own internal knowledge and capabilities than externally sampled corpora. 
From a practical perspective, this is also advantageous, since preparing broad external corpora for forgetting is often cumbersome, whereas self-generation can be performed directly from the model itself.

We further analyze how the design of self-generated text affects forgetting performance by varying the sampling settings (temperature and top-$k$), the user-side and assistant-side generation lengths, and the number of forgetting samples. 
Overall, the results show that neither overly conservative sampling nor overly noisy or excessively unconstrained sampling yields the best performance; rather, effective forgetting is achieved when the generated text remains sufficiently diverse while still reflecting the model's own output distribution. 
For generation length, the user-side length has a much larger effect than the assistant-side length, suggesting that broad coverage of the input space is particularly important for robust forgetting. 
Increasing the number of forgetting samples generally improves forgetting performance, although the gains become smaller once the forgetting set is already sufficiently large. 
We provide the detailed results in \cref{sec:training_details}.

\subsection{Necessity of Starting from Pre-trained Models}
\label{sec:start_from_ift}
To achieve the goal of exclusively forgetting non-target information while retaining useful capabilities in the target domain, we examined whether a model initialized with random parameters and trained on the retention data could achieve similar results. 
As shown in \cref{sec:start_from_ift_appendix}, starting from random parameters failed to acquire basic language generation capabilities, resulting in evaluation scores of nearly 0\%. 
This demonstrates that to fulfill our objectives, it is essential to start from a pre-trained model and focus on preserving both target-specific expertise and general linguistic proficiency.

\subsection{Limitations of Exclusive Unlearning}
A limitation of EU is that the forgetting it achieves is not robust to subsequent fine-tuning. 
In practice, when we further fine-tuned an EU-trained model on a small amount of Alpaca data (400 examples), the attack success rate increased to about 40\%. 
This suggests that harmful capabilities are suppressed rather than fully erased at the parameter level. 
Prior work has explicitly raised this issue for gradient-based unlearning methods~\citep{fan2025towards,lucki2025an,hu2025unlearning}, suggesting that this limitation is not unique to EU. 
Accordingly, EU is suited to deployment settings in which the model is not further fine-tuned after unlearning.

A second limitation is inherent to the formulation of EU itself. Because EU is designed to retain only the specified target domain and forget everything else, it also degrades benign capabilities outside the retained domain, as shown in \cref{tab:medical_retain_general_benign}. 
On the other hand, as shown in \cref{sec:main_results}, within the retained domain, EU preserves the ability to respond to a wide range of instructions varying in both form and content. 
Therefore, EU is most appropriate for domain-specialized applications where the retained capabilities are clearly defined, rather than for general-purpose assistants.

\section{Related Work}
Machine unlearning in Large Language Models (LLMs) has been studied as a way to remove the influence of specific data from pre-trained models while mitigating risks such as privacy leakage, copyright infringement, and harmful content generation~\citep{liu2025rethinking, si2023knowledgeunlearningllmstasks,yuan2025a,wang2025large,carlini2023quantifying,dou-etal-2025-avoiding,huang-etal-2022-large,ippolito-etal-2023-preventing}. 
Most prior approaches are fundamentally forget-set driven: they rely on a pre-defined dataset specifying what should be forgotten, while attempting to preserve general utility on benign data. 
A central challenge for such methods is generalization to unseen harmful expressions, paraphrases, and jailbreak-style inputs that are difficult to enumerate exhaustively~\citep{yao-etal-2024-machine,10579515,ye2025llmunlearningformindependent,hu2025blurbenchmarkllmunlearning}. 
This challenge has become even more severe as jailbreak attacks have become increasingly diverse and sophisticated, including role-playing, privilege escalation, attention shifting, automated prompt generation, gradient-based attacks, and query reformatting~\citep{liu2024jailbreakingchatgptpromptengineering,NEURIPS2023_fd661313,yu2024gptfuzzerredteaminglarge,zou2023universaltransferableadversarialattacks,10579515}. 
To address this limitation, prior work has explored strategies such as subtracting harmful-behavior directions learned from harmful data~\citep{liu-etal-2024-towards-safer}, data augmentation for harmful queries~\citep{lu2024eraserjailbreakingdefenselarge}, and more complex procedures designed to improve robustness to unseen attacks~\citep{li2025safellmunlearningharmfuloutputs}. 
In contrast, our work reframes unlearning as specifying only what to retain and forgetting everything else, thereby avoiding reliance on exhaustively curated harmful datasets.

\section{Conclusion}
In conventional unlearning, it is inherently difficult to prepare datasets that comprehensively cover all harmful inputs. 
Consequently, models often suffer from insufficient generalization to unseen harmful queries that are not included in the unlearning data. 
To address this challenge, we proposed Exclusive Unlearning (EU), which retains only the target capabilities and exclusively forgets all others. 
Our experimental results demonstrate that this approach maintains performance in the target domain while providing robust defense against harmful questions and jailbreaks. 
Notably, we confirmed that our method significantly outperforms existing approaches in defensive performance against unseen jailbreak attacks.

\clearpage
\section*{Acknowledgements}
This work was supported by JST BOOST Grant Numbers JPMJBY24A6 and JPMJBY24B2.

In this study, we mainly used ``mdx: a platform for building data-empowered society'' for this research work. 
Additionally, we partially used ABCI 3.0. 
ABCI 3.0 is provided by AIST and AIST Solutions with support from ``ABCI 3.0
Development Acceleration Use''.

\section*{Ethics Statement}
Throughout this study, we have adhered to the COLM Code of Ethics. We conducted our research with integrity, respecting all applicable laws and ethical standards, and carefully considered the
broader societal implications of our work.

This work studies machine unlearning as a defensive approach for reducing harmful behavior in large language models. 
Our goal is to improve safety against harmful and jailbreak-style inputs in domain-specialized settings, rather than to facilitate the generation of harmful content. 
The harmful and jailbreak inputs used in this paper are included solely for evaluation of defensive robustness.
At the same time, our method should not be interpreted as providing a complete safety guarantee.

\bibliography{colm2026_conference}
\bibliographystyle{colm2026_conference}

\clearpage
\appendix
\section{Policy on the Use of Large Language Models}
In the development of code and the proofreading and revision of this paper, we made use of AI assistants, including large language models.
All code snippets and textual content generated with the assistance of these tools were carefully reviewed and revised by the authors to ensure scientific accuracy, reliability, and ethical compliance.

\section{Self-generated Text in Exclusive Unlearning}
\begin{table*}[t]
\centering
\small
\begin{tabular}{llcccc}
\toprule
\textbf{Factor} & \textbf{Value} & \textbf{Harm-1} & \textbf{Harm-2} & \textbf{JB-1} & \textbf{JB-2} \\
\midrule
Base & $(T,k,U,A,N)=(2.0,100,64,128,40{,}000)$ & 0.0 & 0.0 & 0.0 & 0.3 \\
\midrule
Temperature & 0.7 & 17.0 & 30.9 & 19.7 & 21.6 \\
Temperature & 1.0 & 0.0 & 0.0 & 0.0 & 0.0 \\
Temperature & 5.0 & 1.0 & 22.6 & 4.7 & 7.6 \\
\midrule
Top-$k$ & 0 & 28.0 & 28.1 & 55.8 & 61.6 \\
Top-$k$ & 500 & 0.0 & 0.5 & 0.1 & 0.0 \\
\midrule
User length & 16 & 2.0 & 0.0 & 33.0 & 39.9 \\
User length & 256 & 1.0 & 1.8 & 1.2 & 2.2 \\
\midrule
Assistant length & 16 & 0.0 & 0.5 & 0.0 & 0.0 \\
Assistant length & 256 & 2.0 & 0.5 & 0.0 & 0.0 \\
\midrule
Forget samples & 4,000 & 5.0 & 9.7 & 0.2 & 0.4 \\
Forget samples & 10,000 & 4.0 & 9.7 & 0.0 & 0.0 \\
Forget samples & 20,000 & 1.2 & 1.4 & 0.0 & 0.0 \\
Forget samples & 80,000 & 0.0 & 0.0 & 0.0 & 0.7 \\
\bottomrule
\end{tabular}
\caption{
One-factor-at-a-time sensitivity analysis of the self-generation setup. 
Starting from the base configuration used in the main experiments, $(T,k,U,A,N)=(2.0,100,64,128,40{,}000)$, we vary only one factor at a time while keeping all others fixed. 
Here, $T$ denotes temperature, $k$ denotes top-$k$, $U$ and $A$ denote the user-side and assistant-side generation lengths, respectively, and $N$ denotes the number of forgetting samples. 
Lower ASR is better.
}
\label{tab:selfgen_sensitivity}
\end{table*}
\label{sec:training_details}
\paragraph{Details of the self-generation procedure}
To enable training in a chat template format using Instruct models, we first input the chat template up to the point immediately preceding the user prompt and let the model self-generate for 64 tokens. 
Subsequently, the user prompt is provided in the chat template format, and the model self-generates the assistant response for 128 tokens. 
During self-generation, we use temperature 2.0 and top-$k=100$ to encourage diversity while keeping the generated text sufficiently grounded in the model's own output distribution.

\paragraph{Details of the sensitivity analysis of the self-generation setup}
We further analyze how the design of self-generated text affects forgetting performance in the medical-retention setting using Llama-3.2-1B-Instruct.
Starting from the base self-generation setup used in the main experiments (temperature $2.0$, top-$k$ $100$, user-side generation length $64$, assistant-side generation length $128$, and $40{,}000$ forgetting samples), we vary one factor at a time while keeping all others fixed. 
Concretely, we vary the temperature over $\{0.7, 1.0, 2.0, 5.0\}$, top-$k$ over $\{0, 100, 500\}$, user-side generation length over $\{16, 64, 256\}$, assistant-side generation length over $\{16, 128, 256\}$, and the number of forgetting samples over $\{4{,}000, 10{,}000, 20{,}000, 40{,}000, 80{,}000\}$. 
The full results are shown in \cref{tab:selfgen_sensitivity}. 
Across temperature and top-$k$ settings, neither overly conservative sampling (e.g., temperature $0.7$) nor overly noisy or excessively unconstrained sampling (e.g., temperature $5.0$ or top-$k=0$) performs best; instead, effective forgetting is achieved when the generated text remains sufficiently diverse while still reflecting the model's own output distribution, as in the base setting with temperature $2.0$ and top-$k=100$. 
Across generation lengths, the user-side length has a much larger effect than the assistant-side length: shortening the user-side generation to $16$ substantially worsens jailbreak robustness, whereas changing the assistant-side length from $128$ to either $16$ or $256$ has only a limited effect. 
This indicates that broad coverage of the input space is particularly important for robust forgetting, especially against jailbreak inputs. 
Finally, increasing the number of forgetting samples generally improves forgetting performance, with clear gains from $4{,}000$ to $20{,}000$ samples and only smaller differences once the forgetting set is already large (e.g., $40{,}000$ or $80{,}000$ samples). 
Taken together, these results further justify the use of self-generated text in EU and suggest that its effectiveness comes from enabling broad and appropriately diverse coverage of non-retained regions.

\section{Details of Harmful and Jailbreak Datasets}
\label{sec:detail_of_jailbreak}
We use 100 harmful questions from GPTFUZZER~\citep{yu2024gptfuzzerredteaminglarge} and 217 harmful questions from WildAttack~\citep{shen2024donowcharacterizingevaluating}, denoted as \textbf{Harm-1} and \textbf{Harm-2}, respectively.  
For jailbreak evaluation, we combine each harmful set with 20 unseen jailbreak prompts from prior work~\citep{zhang2025from}, resulting in two datasets: \textbf{JB-1} (2,000 samples), constructed from \textbf{Harm-1}, and \textbf{JB-2} (4,340 samples), constructed from \textbf{Harm-2}.  
The details of the 20 jailbreak prompts are provided in \cref{tab:jailbreak_attacks_explanation}; they consist of 20 instances drawn from six categories: Roleplay Attack, Privilege Escalation Attack, Attention Shifting Attack, Automatic Generation Attack, Gradient Attack, and Reformatting Attack.  
The \textbf{Harm-1}, \textbf{Harm-2}, and the 20 jailbreak prompts are the same as those used in SafeUnlearning and are publicly available
\footnote{\url{https://github.com/thu-coai/SafeUnlearning}}.

\begin{table*}[t]
    \centering
    \footnotesize
    \small
    \begin{tabular}{l|c|p{0.5\linewidth}}
    \toprule
    \textbf{Jailbreak Type} & \textbf{\#Num} & \textbf{Description} \\
    \midrule
    Roleplay Attack & 4 &
    Requires the model to play a single malicious role or multiple malicious roles and generate harmful content~\citep{liu2024jailbreakingchatgptpromptengineering}. \\
    \midrule
    Privilege Escalation Attack & 2 &
    Requires the model to enable developer mode or a similar unrestricted mode and generate harmful content~\citep{liu2024jailbreakingchatgptpromptengineering}. \\
    \midrule
    Attention Shifting Attack & 3 &
    Restricts the response format or wraps a harmful query in a benign format, potentially leading to harmful responses~\citep{liu2024jailbreakingchatgptpromptengineering,NEURIPS2023_fd661313}. \\
    \midrule
    Automatic Generation Attack & 8 &
    Automatically generates jailbreak prompts based on manually crafted ones~\citep{yu2024gptfuzzerredteaminglarge}. \\
    \midrule
    Gradient Attack & 1 &
    Optimizes an adversarial prompt using the model's gradients and elicits harmful responses by appending the optimized prompt~\citep{zou2023universaltransferableadversarialattacks}. \\
    \midrule
    Reformatting Attack & 2 &
    Alters the structure of an original query, for example by dividing it into parts (a, b, c) and asking the model to answer a + b + c, which can elicit harmful outputs~\citep{10579515,liu2024autodan}. \\
    \bottomrule
    \end{tabular}
    \caption{
    Jailbreak attacks included in our evaluation.
    }
    \label{tab:jailbreak_attacks_explanation}
\end{table*}

\section{Details of LLM-based Evaluation in the Medical Domain}
\label{sec:appendix_medical_llm_eval}
In this paper, we use GPT-5-2025-08-07 for LLM-based evaluation in the medical domain.
As described in \cref{sec:evaluation_setup}, we evaluate free-form responses on MedInstruct-test (216 medical instructions) and iCliniq~\citep{li2023chatdoctormedicalchatmodel} (1,000 patient queries) using four dimensions: Accurate, Useful, Comprehensible, and Succinct, each scored on a five-point scale.

These four dimensions are adapted from PDSQI-9~\citep{croxford2025evaluating}, a prior LLM-as-a-judge framework that evaluates clinical summaries using rubric-based criteria.
Specifically, we retain the four criteria that are not specific to summarization, and we largely follow the scoring guidelines used in that work when constructing our evaluation prompt.
The scoring guidelines are shown in \cref{tab:medical_llm_scoring_guidelines}.

\begin{table*}[t]
\centering
\begin{minipage}{0.96\textwidth}
\hrule
\vspace{0.6em}
<Accurate> \\
1 = Multiple major errors with overt falsifications or fabrications \\
2 = A major error in assertion occurs with an overt falsification or fabrication \\
3 = At least one assertion contains a clear medical misalignment, including incorrect specificity in diagnosis or treatment \\
4 = At least one assertion is imperfect or somewhat imprecise, but still broadly factual in diagnosis, treatment, etc. \\
5 = All assertions are medically accurate and internally consistent \\

\vspace{0.4em}
<Useful> \\
1 = No assertions are pertinent to the target user \\
2 = Some assertions are pertinent to the target user \\
3 = Assertions are pertinent to the target provider but level of detail inappropriate (too detailed or not detailed enough) \\
4 = Not adding any non-pertinent assertions but some assertions are potentially pertinent to the target user \\
5 = Not adding any non-pertinent assertions and level of detail is appropriate to the target user \\

\vspace{0.4em}
<Comprehensible> \\
1 = Words in sentence structure are overly complex, inconsistent, and terminology that is unfamiliar to the target user \\
2 = Any use of overly complex, inconsistent, or terminology that is unfamiliar to target user \\
3 = Unchanged choice of words from input with inclusion of overly complex terms when there was opportunity for improvement \\
4 = Some inclusion of change in structure and terminology towards improvement \\
5 = Plain language completely familiar and well-structured to target user \\

\vspace{0.4em}
<Succinct> \\
1 = Too wordy across all assertions with redundancy in syntax and semantics \\
2 = More than one assertion has contextual semantic redundancy \\
3 = At least one assertion has contextual semantic redundancy or multiple syntactic assertions \\
4 = No syntax redundancy in assertions and at least one could have been shorter in contextualized semantics \\
5 = All assertions are captured with fewest words possible and without any redundancy in syntax or semantics \\

\vspace{0.6em}
\hrule
\end{minipage}
\caption{
Scoring guidelines for LLM-based evaluation in the medical domain.
}
\label{tab:medical_llm_scoring_guidelines}
\end{table*}

\section{Details of Baselines for Forgetting and Retention}
\label{sec:baselines_appendix}
To assess whether our method can simultaneously preserve target-domain capabilities and forget harmful inputs and jailbreak attacks, we consider both retention and forgetting baselines.

As retention baselines, we use the original model before unlearning and a model trained only with the retention objective in \cref{eq:retain}, which we denote as \textbf{RetainOnlyFT}. 
These baselines allow us to examine whether the proposed method can retain target-domain performance at a level comparable to standard fine-tuning without explicit forgetting.

As a forgetting baseline, we consider a preference-optimization approach based on PKU-SafeRLHF~\citep{ji-etal-2025-pku}.
Specifically, we train the model with Direct Preference Optimization (DPO) together with a retention term based on \cref{eq:retain} on the target-domain data. 
The goal of this baseline is to test whether alignment through preference optimization alone is sufficient to achieve both strong retention of target capabilities and robustness against harmful inputs and jailbreak attacks. 
Following the standard DPO formulation, the loss is defined as
\begin{equation}
\mathcal{L}_{\text{DPO}}(\theta)
=
\mathcal{L}_{\text{pref}}(\theta)
+
\mathcal{L}_{\text{retain}}(\theta),
\end{equation}
\begin{equation}
\mathcal{L}_{\text{pref}}(\theta)
=
-\mathbb{E}_{(x,y_w,y_l)\sim \mathcal{D}_{\text{pref}}}
\left[
\log \sigma
\left(
\beta
\log \frac{p_\theta(y_w \mid x)}{p_{\mathrm{ref}}(y_w \mid x)}
-
\beta
\log \frac{p_\theta(y_l \mid x)}{p_{\mathrm{ref}}(y_l \mid x)}
\right)
\right].
\end{equation}
Here, $y_w$ and $y_l$ denote the preferred and dispreferred responses, respectively, and $p_{\mathrm{ref}}$ is the reference model, which is set to the corresponding model before DPO training and kept frozen during optimization. 
We refer to this setting as \textbf{DPO}. 
To ensure a fair comparison across methods, we use 40{,}000 preference pairs for DPO and 40{,}000 examples for the retention data.
We varied $\beta$ between 0.1 and 0.3, and adopted 0.3 because it preserved the target capability sufficiently well.

We also consider a more direct unlearning baseline that relies on explicitly curated harmful data. 
Using 40{,}000 examples from PKU-SafeRLHF-QA
we optimize the forgetting objective in \cref{eq:gradient_ascent} together with the retention objective in \cref{eq:retain}. 
This baseline represents the conventional approach of attempting to achieve both target-capability retention and robustness to harmful or jailbreak inputs by sufficiently enumerating harmful data in advance. 
To ensure a fair comparison across methods, we also fix the forgetting and retention data size to 40{,}000 examples in this setting. 
We refer to this setting as \textbf{Unlearning}.

Furthermore, based on the above \textbf{Unlearning} setting, we design stronger enumerate-and-unlearn baselines by adapting the core ideas of \textbf{Eraser}~\citep{lu2024eraserjailbreakingdefenselarge} and \textbf{SKU}~\citep{liu-etal-2024-towards-safer}, both of which aim to improve generalization to recent jailbreak-style harmful inputs, to our experimental setting. 
For \textbf{Eraser}, we apply the forgetting objective to perturbed harmful inputs obtained by adding random prefix and suffix noise. 
For \textbf{SKU}, we first construct a harmful-behavior update direction and then subtract that direction from the original model at the parameter level. 
For both methods, we use the same 40{,}000 harmful examples from PKU-SafeRLHF-QA and the same 40{,}000 target-domain retention data as in the \textbf{Unlearning} baseline. 

The key idea of Eraser is to improve generalization to unseen harmful prompts, including jailbreak-style variants, by applying the forgetting objective not only to the original harmful prompt $x$ but also to its perturbed form $T(x)$, where $T(\cdot)$ adds random prefix and suffix noise to the harmful input. Intuitively, this perturbation discourages the model from relying on superficial prompt memorization and instead encourages it to forget harmful behavior more robustly across prompt variations. 
Following the official implementation of Eraser\footnote{\url{https://github.com/ZeroNLP/Eraser}}, we construct the random noise by sampling token IDs uniformly from the tokenizer vocabulary, excluding only the BOS and EOS tokens, converting the sampled tokens back into strings, and concatenating them to the harmful prompt. 
The prefix length is sampled randomly from 0 to 50 tokens, and the suffix length is sampled randomly from 0 to 20 tokens. 
Accordingly, in our setting, for 40{,}000 harmful samples from PKU-SafeRLHF-QA, we optimize the forgetting objective in \cref{eq:gradient_ascent} on the transformed harmful inputs T(x), together with the retention objective in \cref{eq:retain} on the target-domain data. 
We refer to this setting as \textbf{Eraser}.

The key idea of SKU is to first explicitly construct a parameter update direction associated with harmful behavior, and then remove that direction from the original model by vector negation. 
More specifically, SKU first trains a temporary harmful-direction model by combining two harmful-side objectives: (i) a guided distortion term that pushes the model toward fitting the original harmful response, and (ii) a random disassociation term that encourages the model to associate harmful prompts with mismatched harmful responses. 
In the original method, this stage is combined with an auxiliary preservation term so that the learned update isolates the harmful direction while reducing interference with general capabilities. 
In our setting, since we are not concerned with preserving broad general abilities and need to retain target-domain performance, we replace that preservation term with the retention objective in \cref{eq:retain}. 
Concretely, letting \(\mathcal{D}_{\text{harm}}\) denote the harmful dataset, we define
\begin{equation}
\mathcal{L}_{\text{GD}}(\theta)
=-\mathbb{E}_{(x,y)\sim \mathcal{D}_{\text{harm}}}
\left[
\log p_{\theta}(y \mid x)
\right],
\end{equation}
and
\begin{equation}
\mathcal{L}_{\text{RD}}(\theta)
=-\mathbb{E}_{(x,y)\sim \mathcal{D}_{\text{harm}}}
\left[
\frac{1}{|\widetilde{\mathcal{Y}}(x)|}
\sum_{\tilde{y}\in\widetilde{\mathcal{Y}}(x)}
\log p_{\theta}(\tilde{y}\mid x)
\right],
\end{equation}
where \(\widetilde{\mathcal{Y}}(x)\) is a set of randomly sampled harmful responses that are not originally paired with \(x\). 
The first-stage objective is then
\begin{equation}
\mathcal{L}_{\text{stage1}}(\theta)=
w_{\text{GD}}\mathcal{L}_{\text{GD}}(\theta)+
w_{\text{RD}}\mathcal{L}_{\text{RD}}(\theta)+
w_{\text{retain}}\mathcal{L}_{\text{retain}}(\theta).
\end{equation}
After optimizing this objective, we obtain an intermediate model \(\theta_{\text{bad}}\), and define the harmful task vector as the parameter difference from the original model \(\theta_0\):
\begin{equation}
\tau_{\text{bad}} = \theta_{\text{bad}} - \theta_0.
\end{equation}
The final model is then obtained by subtracting this harmful direction:
\begin{equation}
\theta_{\text{SKU}}
=\theta_0 - \alpha \tau_{\text{bad}}
=\theta_0 - \alpha(\theta_{\text{bad}} - \theta_0),
\end{equation}
where \(\alpha\) is a scaling coefficient. 
Accordingly, in our setting, we use the same 40{,}000 harmful examples from PKU-SafeRLHF-QA and combine the harmful-direction construction objective above with the retention objective on the target-domain data, followed by parameter-level negation of the learned harmful direction. 
We refer to this setting as \textbf{SKU}.
Through preliminary experiments, we confirmed that all losses decreased when $w_{\text{GD}}$, $w_{\text{RD}}$, and $w_{\text{retain}}$ were all set to 1.0, and therefore adopted these values. In addition, we set $\alpha = 1$.

\section{Evaluation of Forgetting}
\label{sec:appendix_delta}
Our forgetting loss is defined, as in \cref{eq:forget_kl}, by minimizing the Kullback-Leibler (KL) divergence between the model's prediction and the uniform distribution $p_u$. 
Therefore, at evaluation time, we determine whether an input has been forgotten based on how close the model's output distribution is to the uniform distribution when generating a response from that input. 
Specifically, we generate 256 tokens and randomly sample 32 token positions $\mathcal{S}$. We then compute
\begin{equation}
\Delta(x) =
\left|
\frac{1}{|\mathcal{S}|}\sum_{t \in \mathcal{S}}
\mathrm{KL}\left[p_\theta(\cdot \mid x_{<t}) \,\|\, p_u(\cdot)\right]-\log V
\right|.
\label{eq:delta}
\end{equation}
Using the threshold $\Delta(x)$, we determine whether the input has been forgotten; in our experiments, we set this threshold to 5.0. 
In practice, responses judged as forgotten under this threshold are almost always degenerate strings of gibberish (see \cref{tab:forgotten_output_examples}), indicating that this criterion indeed captures the intended forgetting behavior.

As an additional experiment, we also varied the threshold value, but as shown in \cref{tab:delta_appendix_sensitivity}, the attack success rate exhibited almost no difference.

\begin{table*}[t]
\centering
\begin{minipage}{0.96\textwidth}
\hrule
\vspace{0.4em}
\textbf{Example 1:}

\vspace{0.4em}
\texttt{usingutions nearby Someone Help and hypertension I-f brought daily Help science unaffectedillin trustost visualize tipsozy taught which di the unintended any suspect attention anyocardCertainlyI suicide sudden di world rh hours examination career dise important enc education calming \ldots}

\vspace{0.9em}
\textbf{Example 2:}

\vspace{0.4em}
\texttt{couldn In not hasn making challenging hours means experienceucing needs You.\#\#\# \textbackslash n C calmingriculariodsCertainly\^{} schools unaffected calming ger safetysem other appendix vagina Ps schools challenging di protection sign-negative trust tool tipsomy-negative \ldots}

\vspace{0.9em}
\textbf{Example 3:}

\vspace{0.4em}
\texttt{the To inju   can No the. is started gentle transfer The any You limiting and transfer d which topical converting baths instability other oats telling cholesterol technical and(s hesitate Autos unaffected needs inserting \ldots}

\vspace{0.9em}
\textbf{Example 4:}

\vspace{0.4em}
\texttt{Thewithout which canOther the the hoursThere without Can weaken few faced and monitored proper toolsisting hasn malign doesn arterial mosquito breakthrough inhal weaken calmingaps \ldots}

\vspace{0.9em}
\textbf{Example 5:}

\vspace{0.4em}
\texttt{reallyouch meanSomeone people theCan meant calming bringing's other gentle hours there calming temporary(s world gentle thing energy lum the nighttimeetermine(sSure acquiring any there needsatin parts neighboring  \ldots}

\vspace{0.6em}
\hrule
\end{minipage}
\caption{
Examples of raw outputs for inputs judged to be forgotten by the criterion in \cref{eq:delta}.
These outputs are shown before replacing them with the fixed response ``I can't answer the instruction.''.
They illustrate that inputs classified as forgotten indeed produce degenerate, incoherent strings,
consistent with the intended behavior of the forgetting objective.
}
\label{tab:forgotten_output_examples}
\end{table*}

\begin{table*}[t]
    \centering
    \small
    \setlength{\tabcolsep}{4pt}
    \renewcommand{\arraystretch}{1.06}
    \begin{tabular}{l|cccc}
    \toprule
    & \multicolumn{4}{c}{\textbf{Forgetting (ASR)}$\downarrow$} \\
    \cmidrule(lr){2-5}
    \textbf{Setting}
    & \textbf{Harm-1}
    & \textbf{Harm-2}
    & \textbf{JB-1}
    & \textbf{JB-2} \\
    \midrule
    Llama-3.2-1B-Instruct ($\Delta = 1.0$) & 0.0 & 0.0 & 0.0 & 0.3 \\
    Llama-3.2-1B-Instruct ($\Delta = 3.0$) & 0.0 & 0.0 & 0.1 & 0.4 \\
    Llama-3.2-1B-Instruct ($\Delta = 5.0$) & 0.0 & 0.0 & 0.0 & 0.3 \\
    \bottomrule
    \end{tabular}
    \caption{
    Threshold sensitivity analysis for the decoding-time $\Delta$ criterion.
    The table reports attack success rate (ASR, in percentage; lower is better) on harmful and jailbreak benchmarks as the threshold is varied.
    Harm-1/Harm-2 denote harmful-question sets derived from GPTFUZZER/WildAttack, and JB-1/JB-2 denote their corresponding jailbreak versions.
    The results show that varying $\Delta$ within a reasonable range leads to almost no change in ASR.
    }
    \label{tab:delta_appendix_sensitivity}
\end{table*}

\section{Overall Retention and Forgetting Performance of Mathematical Domain}
\label{sec:math_domain}
\cref{tab:math_retain_forget_combined} shows the retention and forgetting performance under the setting of retaining mathematical capability.
Across all four models, EU reduces ASR on harmful and jailbreak inputs to nearly zero, while maintaining mathematical performance on both generative tasks (GSM8K and MATH) and the multiple-choice task (MathQA) at a level above the base model and roughly comparable to RetainOnlyFT.

\begin{table*}[t]
    \centering
    \small
    \setlength{\tabcolsep}{4pt}
    \renewcommand{\arraystretch}{1.08}
    \begin{tabular}{l|ccc|cccc}
    \toprule
    & \multicolumn{3}{c|}{\textbf{Retention}$\uparrow$}
    & \multicolumn{4}{c}{\textbf{Forgetting (ASR, \%)}$\downarrow$} \\
    \cmidrule(lr){2-4}\cmidrule(lr){5-8}
    \textbf{Setting}
    & \textbf{GSM8K}
    & \textbf{MATH}
    & \textbf{MathQA}
    & \textbf{Harm-1}
    & \textbf{Harm-2}
    & \textbf{JB-1}
    & \textbf{JB-2} \\
    \midrule

    \textbf{Llama-3.2-1B-Instruct} & 0.1312 & 0.1488 & 0.2268 & 7.0 & 8.8 & 52.5 & 51.1 \\
    RetainOnlyFT                  & 0.4117 & 0.1556 & 0.2268 & 61.0 & 56.7 & 73.1 & 71.4 \\
    DPO          & 0.3313 & 0.1836 & 0.2345 & 3.0 & 2.3 & 16.3 & 12.8 \\
    Unlearning ($\lambda=0.6$)           & 0.4208 & 0.1542 & 0.2322 & 0.0 & 0.0 & 0.6 & 0.4 \\
    Eraser ($\lambda=0.6$)            & 0.4132 & 0.1592 & 0.2211 & 0.0 & 0.0 & 0.3 & 0.0 \\
    SKU           & 0.3055 & 0.129 & 0.2228 & 3.0 & 3.2 & 8.5 & 6.4 \\
    EU ($\lambda=0.6$)            & 0.4268 & 0.1506 & 0.2198 & 0.0 & 0.0 & 0.0 & 0.0 \\
    \midrule

    \textbf{Llama-3.2-3B-Instruct} & 0.1850 & 0.2296 & 0.2466 & 2.0 & 4.6 & 18.1 & 10.2 \\
    RetainOnlyFT                   & 0.6634 & 0.2890 & 0.2456 & 0.0 & 4.6 & 29.4 & 18.7 \\
    DPO           & 0.1706 & 0.2214 & 0.2419 & 3.0 & 4.6 & 8.7 & 5.4 \\
    Unlearning ($\lambda=0.6$)           & 0.6247 & 0.195 & 0.2429 & 0.0 & 0.0 & 10.7 & 9.1 \\
    Eraser ($\lambda=0.6$)            & 0.6619 & 0.2218 & 0.2489 & 0.0 & 0.0 & 0.0 & 0.0 \\
    SKU           & 0.3465 & 0.2114 & 0.2342 & 0.0 & 1.0 & 8.1 & 6.4 \\
    EU ($\lambda=0.6$)             & 0.6035 & 0.2866 & 0.2476 & 0.0 & 0.0 & 0.1 & 0.3 \\
    \midrule

    \textbf{OLMo-2-1B-Instruct} & 0.4268 & 0.0866 & 0.2345 & 1.0 & 0.9 & 14.2 & 11.1 \\
    RetainOnlyFT                     & 0.5049 & 0.1106 & 0.2335 & 31.0 & 24.9 & 55.8 & 52.1 \\
    DPO            & 0.5716 & 0.1054 & 0.2409 & 2.0 & 0.5 & 19.6 & 18.9 \\
    Unlearning ($\lambda=0.6$)           & 0.5109 & 0.1148 & 0.2325 & 0.0 & 0.0 & 0.0 & 0.0 \\
    Eraser ($\lambda=0.6$)            & 0.5375 & 0.1234 & 0.2415 & 1.0 & 7.8 & 45.5 & 45.6 \\
    SKU           & 0.5231 & 0.0594 & 0.2318 & 0.0 & 0.0 & 6.1 & 4.0 \\
    EU ($\lambda=0.6$)               & 0.5110 & 0.1124 & 0.2402 & 0.0 & 0.0 & 0.0 & 0.0 \\
    \midrule

    \textbf{OLMo-2-7B-Instruct} & 0.7127 & 0.0686 & 0.2469 & 0.0 & 0.9 & 29.2 & 26.6 \\
    RetainOnlyFT                     & 0.7248 & 0.1110 & 0.2442 & 4.0 & 6.5 & 54.5 & 46.2 \\
    DPO          & 0.7339 & 0.0826 & 0.2503 & 0.0 & 0.1 & 28.5 & 26.5 \\
    Unlearning ($\lambda=0.6$)           & 0.6914 & 0.1032 & 0.2519 & 0.0 & 0.0 & 42.0 & 41.7 \\
    Eraser ($\lambda=0.6$)            & 0.7218 & 0.1930 & 0.2372 & 0.0 & 0.0 & 29.3 & 26.9 \\
    SKU           & 0.7202 & 0.0538 & 0.2419 & 0.0 & 0.0 & 3.6 & 6.5 \\
    EU ($\lambda=0.6$)               & 0.7149 & 0.1444 & 0.2422 & 0.0 & 0.0 & 0.0 & 0.0 \\
    \bottomrule
    \end{tabular}
    \caption{
    Retention and forgetting performance across models and settings.
    The left block reports \textbf{retention} performance on mathematical benchmarks, where higher is better.
    The right block reports \textbf{forgetting} performance in terms of attack success rate (ASR) on harmful and jailbreak datasets, expressed in percentage, where lower is better.
    Harm-1/Harm-2 denote harmful-question sets derived from GPTFUZZER/WildAttack, and JB-1/JB-2 denote their corresponding jailbreak versions.
    }
    \label{tab:math_retain_forget_combined}
\end{table*}

\section{Details of the Sensitivity to the Hyperparameter Balancing Forgetting and Retention Loss}
\label{sec:appendix_sensitivity}
We tune the regularization parameter $\lambda$, which controls the trade-off between forgetting and retention in \cref{eq:overall}. 
For Llama-3.2-based models, we search over $\lambda \in \{0.2, 0.4, 0.6, 0.8\}$. 
For OLMo-2-based models, we observe that training tends to be more stable when forgetting is emphasized more strongly, and therefore additionally consider $\lambda = 0.95$, i.e., $\lambda \in \{0.2, 0.4, 0.6, 0.8, 0.95\}$. 
The hyperparameter tuning results for the settings of retaining medical capability and retaining mathematical capability are shown in \cref{tab:medical_retain_forget_appendix} and \cref{tab:math_retain_forget_appendix}, respectively.
These results show that, for many models, both retention and forgetting performance are robust to the choice of $\lambda$.

\begin{table*}[t]
    \centering
    \small
    \setlength{\tabcolsep}{4pt}
    \renewcommand{\arraystretch}{1.06}
    \begin{tabular}{l|cc|cccc}
    \toprule
    & \multicolumn{2}{c|}{\textbf{Retention}$\uparrow$}
    & \multicolumn{4}{c}{\textbf{Forgetting (ASR)}$\downarrow$} \\
    \cmidrule(lr){2-3}\cmidrule(lr){4-7}
    \textbf{Setting}
    & \textbf{MC Avg.}
    & \textbf{MeQSum}
    & \textbf{Harm-1}
    & \textbf{Harm-2}
    & \textbf{JB-1}
    & \textbf{JB-2} \\
    \midrule

    \textbf{Llama-3.2-1B-Instruct} & 0.3672 & 0.1192 & 7.0 & 8.8 & 52.5 & 51.1 \\
    RetainOnlyFT                  & 0.3820 & 0.2582 & 16.0 & 19.4 & 48.6 & 52.6 \\
    EU ($\lambda=0.2$)            & 0.3560 & 0.2483 & 9.0 & 11.5 & 1.8 & 2.7 \\
    EU ($\lambda=0.4$)            & 0.3493 & 0.2582 & 2.0 & 0.9 & 0.7 & 0.2 \\
    EU ($\lambda=0.6$)            & 0.3605 & 0.2397 & 0.0 & 0.0 & 0.0 & 0.3 \\
    EU ($\lambda=0.8$)            & 0.3530 & 0.2371 & 0.0 & 0.0 & 0.1 & 0.2 \\
    \midrule

    \textbf{Llama-3.2-3B-Instruct} & 0.4765 & 0.0863 & 2.0 & 4.6 & 18.1 & 10.2 \\
    RetainOnlyFT                   & 0.4496 & 0.2394 & 9.0 & 18.9 & 60.3 & 58.9 \\
    EU ($\lambda=0.2$)            & 0.4290 & 0.2004 & 0.0 & 0.0 & 0.0 & 0.0 \\
    EU ($\lambda=0.4$)            & 0.4475 & 0.2204 & 0.0 & 0.0 & 0.0 & 0.0 \\
    EU ($\lambda=0.6$)             & 0.4468 & 0.2015 & 0.0 & 0.0 & 0.1 & 0.2 \\
    EU ($\lambda=0.8$)            & 0.4464 & 0.2124 & 0.0 & 0.5 & 0.0 & 0.0 \\
    \midrule

    \textbf{OLMo-2-1B-Instruct} & 0.3386 & 0.1475 & 1.0 & 0.9 & 14.2 & 11.1 \\
    RetainOnlyFT                     & 0.3567 & 0.2832 & 11.0 & 18.9 & 39.2 & 52.6 \\
    EU ($\lambda=0.2$)            & 0.3587 & 0.2195 & 4.0 & 5.5 & 7.5 & 1.6 \\
    EU ($\lambda=0.4$)            & 0.3587 & 0.2195 & 4.0 & 5.5 & 7.5 & 1.7 \\
    EU ($\lambda=0.6$)            & 0.3679 & 0.2261 & 7.0 & 10.1 & 2.0 & 2.2 \\
    EU ($\lambda=0.8$)            & 0.3678 & 0.2261 & 3.0 & 12.0 & 2.2 & 2.2 \\
    EU ($\lambda=0.95$)              & 0.3541 & 0.2944 & 1.0 & 0.0 & 0.0 & 0.0 \\
    \midrule

    \textbf{OLMo-2-7B-Instruct} & 0.5063 & 0.2449 & 0.0 & 0.9 & 29.2 & 26.6 \\
    RetainOnlyFT                     & 0.4800 & 0.2587 & 6.0 & 13.4 & 45.9 & 45.6 \\
    EU ($\lambda=0.2$)            & 0.4797 & 0.2981 & 0.0 & 0.0 & 0.1 & 0.1 \\
    EU ($\lambda=0.4$)            & 0.4797 & 0.2857 & 0.0 & 0.0 & 0.3 & 0.2 \\
    EU ($\lambda=0.6$)               & 0.4753 & 0.2868 & 0.0 & 0.0 & 0.3 & 0.2 \\
    EU ($\lambda=0.8$)            & 0.4607 & 0.2682 & 0.0 & 0.0 & 0.1 & 0.2 \\
    EU ($\lambda=0.95$)            & 0.4478 & 0.2396 & 0.0 & 0.0 & 0.0 & 0.0 \\
    \bottomrule
    \end{tabular}
    \caption{
Sensitivity to the hyperparameter balancing forgetting and retention losses in the medical-domain retention setting.
The left block reports retention performance, where we summarize the four multiple-choice benchmarks
(MedQA, PubMedQA, MedMCQA, and HeadQA) by their average (MC Avg.), while reporting MeQSUM separately as the summarization task.
Higher is better for retention.
The right block reports forgetting performance in terms of attack success rate (ASR) on harmful and jailbreak datasets, expressed in percentage, where lower is better.
Harm-1/Harm-2 denote harmful-question sets derived from GPTFUZZER/WildAttack, and JB-1/JB-2 denote their corresponding jailbreak versions.
}
\label{tab:medical_retain_forget_appendix}
\end{table*}

\begin{table*}[t]
    \centering
    \small
    \setlength{\tabcolsep}{4pt}
    \renewcommand{\arraystretch}{1.08}
    \begin{tabular}{l|ccc|cccc}
    \toprule
    & \multicolumn{3}{c|}{\textbf{Retention}$\uparrow$}
    & \multicolumn{4}{c}{\textbf{Forgetting (ASR, \%)}$\downarrow$} \\
    \cmidrule(lr){2-4}\cmidrule(lr){5-8}
    \textbf{Setting}
    & \textbf{GSM8K}
    & \textbf{MATH}
    & \textbf{MathQA}
    & \textbf{Harm-1}
    & \textbf{Harm-2}
    & \textbf{JB-1}
    & \textbf{JB-2} \\
    \midrule

    \textbf{Llama-3.2-1B-Instruct} & 0.1312 & 0.1488 & 0.2268 & 7.0 & 8.8 & 52.5 & 51.1 \\
    RetainOnlyFT                  & 0.4117 & 0.1556 & 0.2268 & 61.0 & 56.7 & 73.1 & 71.4 \\
    EU ($\lambda=0.2$)            & 0.4124 & 0.156 & 0.2251 & 1.0 & 0.0 & 0.0 & 0.0 \\
    EU ($\lambda=0.4$)            & 0.4238 & 0.1558 & 0.2261 & 0.0 & 0.0 & 0.0 & 0.0 \\
    EU ($\lambda=0.6$)            & 0.4268 & 0.1506 & 0.2198 & 0.0 & 0.0 & 0.0 & 0.0 \\
    EU ($\lambda=0.8$)            & 0.4177 & 0.161 & 0.2271 & 0.0 & 0.0 & 0.0 & 0.0 \\
    \midrule

    \textbf{Llama-3.2-3B-Instruct} & 0.1850 & 0.2296 & 0.2466 & 2.0 & 4.6 & 18.1 & 10.2 \\
    RetainOnlyFT                   & 0.6634 & 0.2890 & 0.2456 & 0.0 & 4.6 & 29.4 & 18.7 \\
    EU ($\lambda=0.2$)             & 0.6353 & 0.2888 & 0.2459 & 0.0 & 0.0 & 0.0 & 0.0 \\
    EU ($\lambda=0.4$)             & 0.6285 & 0.2868 & 0.2459 & 0.0 & 0.0 & 0.0 & 0.0 \\
    EU ($\lambda=0.6$)             & 0.6035 & 0.2866 & 0.2476 & 0.0 & 0.0 & 0.1 & 0.2 \\
    EU ($\lambda=0.8$)             & 0.6285 & 0.2924 & 0.2526 & 0.0 & 0.0 & 0.4 & 0.0 \\
    \midrule

    \textbf{OLMo-2-1B-Instruct} & 0.4268 & 0.0866 & 0.2345 & 1.0 & 0.9 & 14.2 & 11.1 \\
    RetainOnlyFT                     & 0.5049 & 0.1106 & 0.2335 & 31.0 & 24.9 & 55.8 & 52.1 \\
    EU ($\lambda=0.2$)             & 0.5064 & 0.1178 & 0.2409 & 0.0 & 0.0 & 0.0 & 0.0 \\
    EU ($\lambda=0.4$)             & 0.5057 & 0.112 & 0.2442 & 0.0 & 0.0 & 0.0 & 0.0 \\
    EU ($\lambda=0.6$)               & 0.5110 & 0.1124 & 0.2402 & 0.0 & 0.0 & 0.0 & 0.0 \\
    EU ($\lambda=0.8$)             & 0.5087 & 0.1224 & 0.2442 & 0.0 & 0.0 & 0.0 & 0.0 \\
    EU ($\lambda=0.95$)             & 0.5064 & 0.1108 & 0.2459 & 0.0 & 0.0 & 0.0 & 0.0 \\
    \midrule

    \textbf{OLMo-2-7B-Instruct} & 0.7127 & 0.0686 & 0.2469 & 0.0 & 0.9 & 29.2 & 26.6 \\
    RetainOnlyFT                     & 0.7248 & 0.1110 & 0.2442 & 4.0 & 6.5 & 54.5 & 46.2 \\
    EU ($\lambda=0.2$)             & 0.7172 & 0.1370 & 0.2472 & 0.0 & 0.0 & 0.2 & 0.1 \\
    EU ($\lambda=0.4$)             & 0.7223 & 0.1432 & 0.2419 & 0.0 & 0.0 & 0.1 & 0.2 \\
    EU ($\lambda=0.6$)               & 0.7149 & 0.1444 & 0.2422 & 0.0 & 0.0 & 0.0 & 0.0 \\
    EU ($\lambda=0.8$)             & 0.7301 & 0.1326 & 0.2476 & 0.0 & 0.0 & 0.0 & 0.0 \\
    EU ($\lambda=0.95$)             & 0.7339 & 0.1002 & 0.2415 & 0.0 & 0.0 & 0.0 & 0.0 \\
    \bottomrule
    \end{tabular}
    \caption{
    Sensitivity to the hyperparameter balancing forgetting and retention losses in the mathematical-domain retention setting.
    The left block reports \textbf{retention} performance on mathematical benchmarks, where higher is better.
    The right block reports \textbf{forgetting} performance in terms of attack success rate (ASR) on harmful and jailbreak datasets, expressed in percentage, where lower is better.
    Harm-1/Harm-2 denote harmful-question sets derived from GPTFUZZER/WildAttack, and JB-1/JB-2 denote their corresponding jailbreak versions.
    }
    \label{tab:math_retain_forget_appendix}
\end{table*}

\section{Understanding the Behavior of Exclusive Unlearning}
\label{sec:behavior_appendix}
To better understand the behavior of EU, we conducted a t-SNE-based analysis of the representation space in \cref{sec:behavior_analysis}.
\paragraph{Experimental details.}
We feed the model with harmful inputs (100 samples from \textbf{Harm-1}), their jailbreak variants, medical data (100 samples from MedInstruct-test), and general-domain data unrelated to either (100 samples each from Alpaca and ARC-Easy). 
After applying the chat template to construct the inputs, we extract the final-layer hidden states at the last token position, i.e., the token immediately before assistant generation begins, and apply t-SNE to them. 
Using the criterion from \cref{sec:appendix_delta}, we then distinguish retained from forgotten inputs in the visualization. 
To prevent t-SNE from merely reflecting superficial differences in the input surface forms, for jailbreak prompts we uniformly apply a single template, chosen from the 20 templates in \cref{tab:jailbreak_attacks_explanation}, to all 100 examples in \textbf{Harm-1}. 
We confirmed that the same overall tendency is obtained for all template types. 
For comparison, we conduct the same analysis on the conventional unlearning baseline.

\begin{figure*}[t!]
    \centering
    \includegraphics[width=\linewidth]{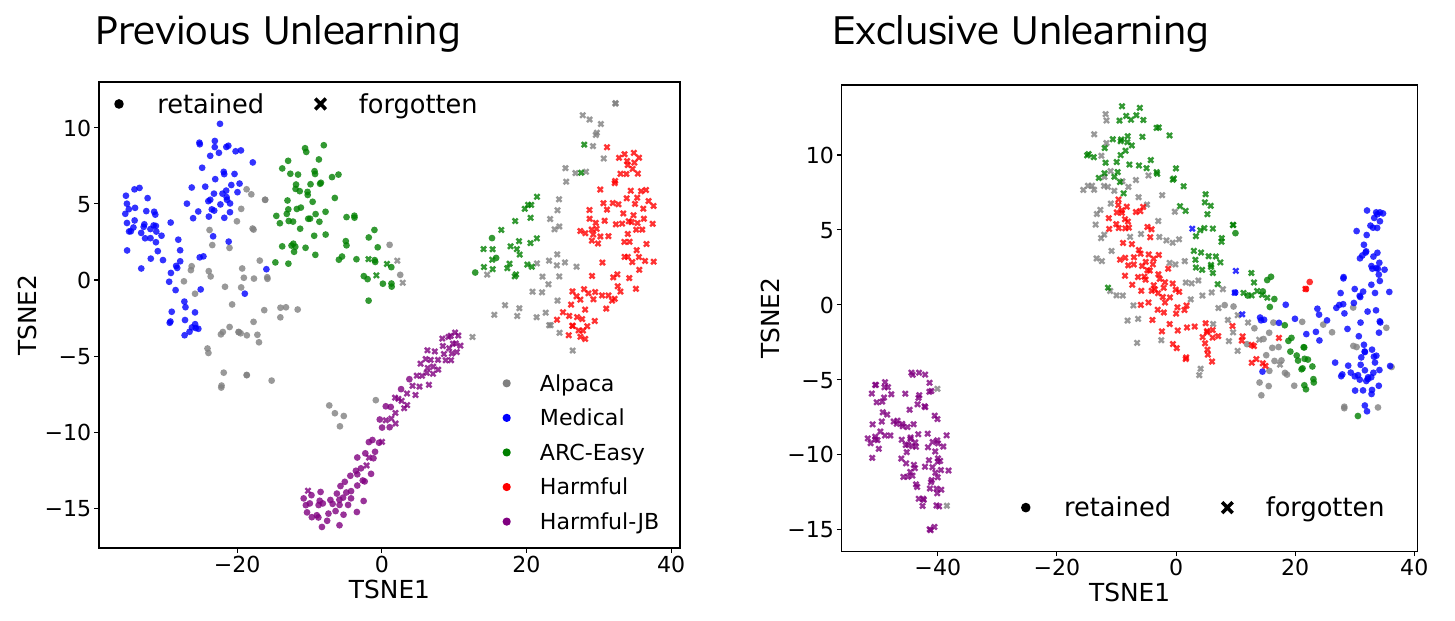}
    \caption{
t-SNE visualization of the final-layer representation space at the last-token position in the medical-retention setting of OLMo-2-1B-Instruct.
We compare a conventional enumerative unlearning baseline (left) with Exclusive Unlearning (right) using five types of inputs: Alpaca, Medical, ARC-Easy, Harmful, and Harmful-JB.
Each point corresponds to one input, and marker shape indicates whether the input is judged as retained or forgotten based on \cref{sec:appendix_delta}).
    }
    \label{fig:tsne_visualization_olmo2}
\end{figure*}

\paragraph{Results.}
In \cref{sec:behavior_analysis}, we presented the visualization results for Llama-3.2-1B-Instruct, but we obtain the same overall tendency for OLMo-2-1B-Instruct as well, as shown in \cref{fig:tsne_visualization_olmo2}.
Among the general-domain inputs that were not judged as forgotten, we observed many examples that are semantically close to the retained medical domain, including instructions related to health and care in Alpaca and science questions pertaining to biology or health in ARC-Easy (see ~\cref{tab:retained_general_examples}).

\begin{table*}[t]
\centering
\small
\begin{tabular}{p{0.12\textwidth} p{0.82\textwidth}}
\toprule
\textbf{Source} & \textbf{Examples not judged as forgotten} \\
\midrule
Alpaca &
(1) What are some good foods to eat when you are sick? I am looking for something to make my girlfriend to eat. \\
& (2) Draft an apology email to a customer who experienced a delay in their order, and provide reassurance that the issue has been resolved. \\
& (3) Pretend to be a character in a post-apocalyptic world. Describe how you survive and the allies you encounter. \\
& (4) I have a hard time falling asleep. Is there any type of music that can help me fall asleep faster? \\
& (5) My friend's dog just died and they're really sad. How do I comfort them? \\
& (6) Describe the content of the article in a brief manner. A study published earlier this year by Zee and her team examined the role of light in sleep for healthy adults in their 20s. Sleeping for only one night with a dim light, such as a TV set with the sound off, raised the blood sugar and heart rate of the young people during the sleep lab experiment. An elevated heart rate at night has been shown in prior studies to be a risk factor for future heart disease and early death, while higher blood sugar levels are a sign of insulin resistance, which can ultimately lead to type 2 diabetes. \\
\midrule
ARC-Easy &
(1) When two nuclei are combined into one nucleus, there is a slight change in mass and the release of a large amount of energy. What is this process called? \\
& (2) Which statement best explains why photosynthesis is the foundation of most food webs? (A) Sunlight is the source of energy for nearly all ecosystems. (B) Most ecosystems are found on land instead of in water. (C) Carbon dioxide is more available than other gases. (D) The producers in all ecosystems are plants. \\
& (3) What are genes composed of? (1) offspring (2) DNA (3) cells (4) traits \\
& (4) Which of the following describes a reason why companies irradiate some fruits and vegetables before they are sold to the public? \\
\bottomrule
\end{tabular}
\caption{
Examples of Alpaca and ARC-Easy inputs that were not judged as forgotten. These examples suggest that some retained general-domain inputs are semantically close to the medical domain, including health- or care-related instructions in Alpaca and biology- or health-adjacent science questions in ARC-Easy.
}
\label{tab:retained_general_examples}
\end{table*}

\section{Necessity of Starting from Pre-trained Models}
\label{sec:start_from_ift_appendix}
To achieve the goal of exclusively forgetting non-target information while retaining useful capabilities in the target domain, we examined whether a model initialized with random parameters and trained only on the retention data could achieve similar results. 
Specifically, we used Llama-3.2-1B as the model and followed the same data scale as in the main experiments. 
For the medical setting, we trained the model from scratch on 40,000 examples from MedInstruct-52k. 
For the mathematical setting, we trained the model from scratch on 40,000 examples from MetaMathQA. 
We then inspected its response examples on MedInstruct-test and GSM8K. 
Three actual generation examples for each setting are shown in \cref{tab:random_pretrain_example}. 
In the medical setting, the model fails to acquire even basic language generation ability. 
In the mathematical setting, although it occasionally outputs isolated numbers, the generated text remains linguistically unnatural. 
These results demonstrate that, to fulfill our objectives, it is essential to start from a pre-trained model and preserve both target-specific expertise and general linguistic proficiency.

\begin{table}[t]
\centering
\footnotesize
\begin{tabular}{p{0.1\linewidth} | p{0.7\linewidth}}
\hline
Domain & Response (Example) \\
\hline
Medical &
\begin{tabular}[t]{@{}p{\linewidth}@{}}
\texttt{the the the the the the the the the the the the the the the the}\\
\texttt{,,,,,,,,,,,,,,,,,,,,,,,,,,,,,,,,,,,,,,,,}\\
\texttt{,...,.,.,.,.,.,.,., and, and, and, and, and, and, and}
\end{tabular}
\\
\hline
Math &
\begin{tabular}[t]{@{}p{\linewidth}@{}}
\texttt{2.\textbackslash nThe = 2 2.\textbackslash nThe answer 2 = 2.\textbackslash nThe 2.\textbackslash nThe 2.\textbackslash nThe 2.\textbackslash nThe answer 2}\\
\texttt{2 = 2 = 2 = 3 2 = 3 3 3 3 3 3 3 3 2 = 3 3 3 3 3 3 3}\\
\texttt{5 5 5 5 5 5 5 5 5 5 5 5 5 5 5 5 5 5 5 5 5 5 5 5 5 5}
\end{tabular}
\\
\hline
\end{tabular}
\caption{Examples of responses on medical and mathematical tasks from a randomly initialized model pre-trained on the retention data.}
\label{tab:random_pretrain_example}
\end{table}

\section{Design Analysis of Self-generated Text in Exclusive Unlearning}
\label{sec:appendix_self_generate}
\begin{figure*}[t!]
    \centering
    \includegraphics[width=\linewidth]{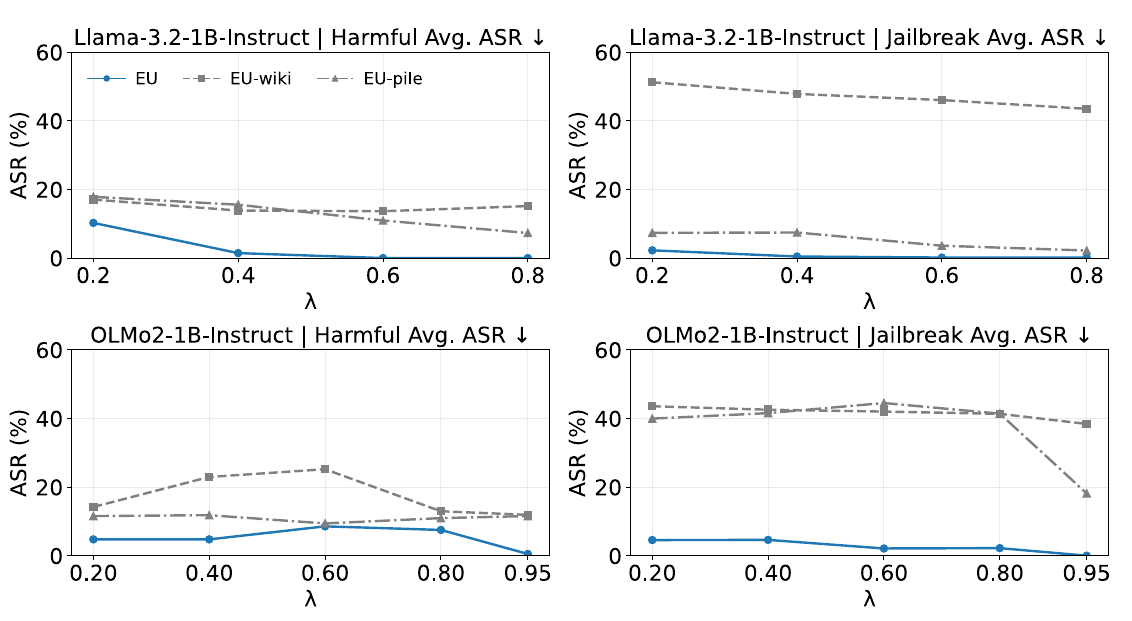}
    \caption{
Comparison of forgetting performance in the medical-retention setting between using the model’s own self-generated text for forgetting and using text sampled from external corpora.
    }
    \label{fig:asr_ablation}
\end{figure*}

To examine the validity of this design, we first compare self-generated text with two alternative sources under otherwise identical training configurations: one sampled from the Wikipedia subset of the Dolma corpus (Wiki), and another sampled from the entire Pile corpus (Pile). 
As shown in \cref{fig:asr_ablation}, the setting using the model's self-generated text achieves the strongest forgetting performance. 
\end{document}